# CCDN: Checkerboard Corner Detection Network for Robust Camera Calibration

Ben Chen, Caihua Xiong*, Qi Zhang

Institute of Rehabilitation and Medical Robotics,
State Key Laboratory of Digital Manufacturing Equipment and Technology,
Huazhong University of Science and Technology, Wuhan, Hubei 430074, China
{benchen,chxiong,edisonzhangqi}@hust.edu.cn

**Abstract.** Aiming to improve the checkerboard corner detection robustness against the images with poor quality, such as lens distortion, extreme poses, and noise, we propose a novel detection algorithm which can maintain high accuracy on inputs under multiply scenarios without any prior knowledge of the checkerboard pattern. This whole algorithm includes a checkerboard corner detection network and some post-processing techniques. The network model is a fully convolutional network with improvements of loss function and learning rate, which can deal with the images of arbitrary size and produce correspondingly-sized output with a corner score on each pixel by efficient inference and learning. Besides, in order to remove the false positives, we employ three post-processing techniques including threshold related to maximum response, non-maximum suppression, and clustering. Evaluations on two different datasets show its superior robustness, accuracy and wide applicability in quantitative comparisons with the state-of-the-art methods, like MATE, ChESS, ROCHADE and OCamCalib.

## 1 Introduction

Camera calibration is a classic task in machine vision with the purpose of estimating the intrinsic parameters as well as the distortion coefficients of camera; the most widely used calibration pattern is the planar checkerboard. Compared with other types of patterns, such as three dimensional objects [1], circles [2] and self-identifying patterns [3], the checkerboard pattern has the stronger robustness with respect to distortion bias and perspective bias [4]. It is also suitable for 3D pose estimation and localization in robot vision, and easy to generate at a low price. However, the checkerboard with poor quality, like low resolution, lens distortion, extreme poses, and sensor noise, can also lead to inaccurate inner corner detection and failed camera calibration.

There are various methods for the checkerboard corner detection. Harris [5], SUSAN [6] and their improved versions [7, 8] adopt distinct corner features to find target points, but those do not generally work well on chessboard pattern. Wang et al. [9] refer to checkerboard corner as the intersection of two adjacent grid lines, which could detect the checkerboard pattern with small lens distortions successfully, but it will be less accurate for wide-angle cameras. ChESS [10] uses the specific features of circular neighborhood around the corners to select the candidates and it is faster and accurate for most cases. However, this method will produce a lot of false detections and heavily depends on the hand-crafted threshold. The widely used checkerboard detection algorithm embedded in OPENCV is based on the work of Vezhnevets [11]. It applies erosion to separate black quadrangles, and then combines them to construct the checkerboard and calculate the inner corners. Rufli et al. extend this algorithm in OCamCalib [12] to be more robust to lens distortion, while for low resolution images and highly distorted images, its detection performance is not as good as ROCHADE [13], a more complex combination of general image features, especially under strong perspective distortion as often presented in wide baseline stereo setups. What's more, these aforementioned algorithms need to know the number of squares of the calibration pattern in advance. Some algorithms attempt to use machine learning methods to detect the corners, such as FAST [14] and FAST-ER [15]. A foray into neural networks is MATE [16], which consists of three convolutional layers to extract the intrinsic feature effectively. But it may cause a little bit more false positives even for medium and high resolution images with little lens distortion.

In this paper we propose a fully convolutional neural network (CNN) model, namely checkerboard corner detection network (CCDN), to find the inner corners of checkerboards under multiply scenarios efficiently. This model can take an image of any size as input and output the response map of corresponding spatial dimensions with a corner score on each pixel. Aided by three post-processing techniques, threshold related to maximum response, non-maximum suppression, and clustering, to eliminate false positive points in different cases, this model is more accurate and robust for the checkerboard corner detection.

The outline of this paper is as follows. Section 2 details the architecture and properties of our checkerboard corners detection network and its training, as well as techniques in the post-processing. The following section describes the datasets for training and testing. Experiments and results are discussed in Section 4. Conclusions are given in the last section.

## 2 Methodology

The whole algorithm can be divided into two parts: the first is a fully convolutional network for extracting a series of corner candidates, which is detailed in Section 2.1; the second part, including threshold related to maximum response, non-maximum suppression and clustering, is described in Section 2.2 to eliminate the false positives.

### 2.1 Checkerboard Corner Detection Network

**Architecture**. A checkerboard corner detection network is presented which can take an image of any size as input and output the response map of corresponding spatial dimensions with a corner score on each pixel. As depicted in Fig. 1, this network consists of six convolutional layers, of which the first and the fourth are followed by max-pooling layers of size $2\times 2$. The ReLU non-linearity [17, 18] is applied to the output of each convolutional layer as the activation function.

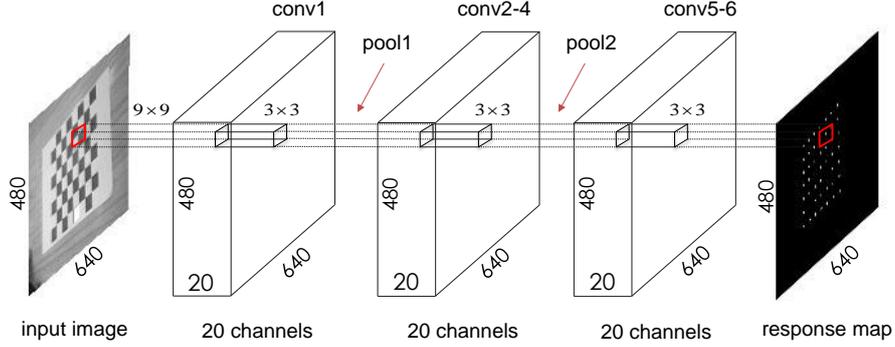

**Fig. 1.** An illustration of the architecture of CCDN. It is a fully convolutional network with six convolutional layers, and the first and fourth are followed by max-pooling layers. The output is a single-channel response map with same size to the input. The activation functions following each convolutional layer are ReLUs.

The kernels of the first layer are intended to extract some useful features from the input image, so the spatial support radius of them should be set large enough to suppress the effect of blur and noise. As research in [16], a larger radius may lose some recall of the real corners while a smaller may falsely detect background pixels as checkerboard corners. To make a tradeoff between recall and precision, here we choose the spatial support radius of four pixels for the first layer, which is shown to be sufficient for corners detection of our model in the result section.

The first convolutional layer filters the input gray-scale image $X$ into 20 channels $L_{1,i}(X)$ with kernels $W_{1,i}$ of size $9\times 9\times 1$ and biases $b_{1,i}$:

$$L_{1,i}(X)(x,y) = \max((W_{1,i}\times X)(x,y)+b_{1,i}, 0). \qquad \forall i=1...20 \qquad (1)$$

The second convolutional layer takes the max-pooled output of the first layer as input and filters it into 20 channels with kernels $W_{2,i,j}$ of size $3\times 3\times 20$ and biases $b_{2,j}$, while the third and fourth convolutional layers of same size filters are followed without any pooling:

$$L_{c,j}(X)(x,y) = \max(\sum_{i=1}^{20}(w_{c,i,j}\times [L_{c-1,i}(X)])(x,y)+b_{c,j}, 0) \quad \forall c=2,3,4;\ j=1...20 \qquad (2)$$

The fifth convolutional layer has 20 kernels of size $3\times 3\times 20$ connected to the max-pooled output of the fourth layer, as explained like Eq. (2). The last convolutional layer combines the 20 channels resulting from the fifth layer into a single response map, with small filters of size $3\times 3\times 20$. The output of this layer is given as:

$$L_6(X)(x,y) = \max(\sum_{i=1}^{20}(W_{6,i} \times [L_{5,i}(X)])(x,y) + b_3, 0) \tag{3}$$

Note that the stride of kernels in both convolutional layers and max-pooling layers is one pixel, and the zero padding is applied to make the output feature map have the equivalent dimensions with input image. This network can be tailored towards application-specific scenarios for its capacity will vary with the depth and settings. We initialize the weights in all convolutional layers from a zero-mean Gaussian distribution with standard deviation 0.1, and the neuron biases with the constant 0.1.

Considering the gray-scale input and the spatial support of $9 \times 9$ for the first filters, our net has 16301 parameters to train, which are a little more than MATE (only 2939 parameters), but much fewer than other types of object detection networks [19-21]. A smaller spatial support can get more effective input samples (291716 for MATE, 296100 for our net, refer to an $640 \times 480$ image), and also little overlap. Furthermore, compared with the convolutional layers in MATE, this net is deeper with more filters, which can extract more features adapted to various scenarios, with no significant increase in time consumption, as well as less risk of overfitting.

**Loss Function**. For training this net, we assign a binary class label of being a corner or not to each input sample: the ground-truth corner locations are assigned a positive label (=1), while those non-corner locations are assigned a negative label (=0), then the corner label of the binary ground-truth image is denoted as $G(x,y)$. All the parameters of the neural net are collected into a single vector $\vec{p}$. Unlike MATE, we use cross entropy instead of mean square error as the loss function, for it is more suitable for discrete output variables [22]. The total loss function is defined as:

$$L(\vec{p}) = \frac{1}{2}\lambda \parallel \vec{p} \parallel_2^2 - \sum_{(x,y)\in\Omega} \begin{cases} \frac{1}{N_p}\log(a(x,y)), & \text{where } G(x,y) = 1, \\ \frac{1}{N_N}\log(1-a(x,y)), & \text{where } G(x,y) = 0. \end{cases} \tag{4}$$

where $a(x,y)$ denotes the clipped output of the last layer as:

$$a(x,y) = \begin{cases} \min(\max(10^{-6}, L_6(X)(x,y)), 1), & \text{where } G(x,y) = 1, \\ \min(\max(0, L_6(X)(x,y)), 1-10^{-6}), & \text{where } G(x,y) = 0. \end{cases} \tag{5}$$

and thus the loss function can be meaningful to all pixels' responses. Noted that there are only few true positives (49, 54, 81, and 156 for our training set) of all effective input samples in an image and the mean loss to each location may make the net mistake all of them for the non-corner locations. In order to eliminate the effect of the disparity that the negative samples are dominate, we normalize the loss by the number of ground-truth positives ($N_p$) and negatives ($N_N$) on each term. In addition, we use the $L^2$ parameter regularization to reduce the net's overfitting. $\lambda$ is a balancing parameter that weights the contribution of regularization term relative to the cross entropy function, and here we set $\lambda=0.01$.

**Training**. This network can be trained by back-propagation and stochastic gradient descent (SGD). We use a batch size of 20 images (about 900 to 3120 positive labels) and a momentum of 0.9 [18]. The learning rate is initially set to 0.01, and then decreases exponentially as the training progresses. The learning rate $v_i$ of $i_{\text{th}}$ iteration is expressed as:

$$v_i = v_0 \sigma^{\lfloor i/\tau \rfloor} \tag{6}$$

where $v_0$ is the initial learning rate, $\sigma$ is the decay rate, and $\tau$ represents the number of iterations required to train all the training images at a time, equal to the total number of training samples divided by the number of those in each batch. $\lfloor i/\tau \rfloor$ with $\lfloor \cdot \rfloor$ denoting floor operation, guarantees the decayed learning rate follows a staircase function so that all samples can be trained with same rate. Appling exponential decay to the learning rate can not only make the net get close to the optimal solution in the early stage of training, but also guarantee that it will not have too many fluctuations in the later stage, so as to get closer to the local optimal solution. Our implementation uses TensorFlow [23].

## 2.2 Techniques for Eliminating the False Positives

The output of this network is a response map with a corner score on each pixel location, and the map is of same spatial dimensions as the input. This section introduces three efficient techniques combined to find the correct checkerboard corners effectively, for they are designed to eliminate false positive points in different cases.

**Threshold Related to Maximum Response**. As the loss function explained above, our model is a binary network for checkerboard corner detection. During the training process it accelerates responses of corner locations closer to 1 and responses of non-corner locations closer to 0. Thus we can set a threshold to distinct them, and the threshold can be adjusted to a higher value (for more precision) or a lower value (for more recall). Furthermore, different images contain different scenes, and their response values are subject to different distributions so a global fixed threshold is not valid for them.

By observing the distribution of the response values, we find that responses of the ground-truth corner locations are often higher than 1, even some false positives may get a value closing to 1, for neither cross entropy or mean square error sets any constrains on the output. This is probably one of the principal reasons why MATE (a fixed threshold of 0.5) is insensitive to the false positives even for pictures with little lens distortion and noise. However, we also find that the number of corner points is approximately linear with the maximum of responses. Here we set half of the maximum as the threshold, which is proved to be useful for most cases.

**Non-maximum Suppression**. After the threshold processing, the locations with lower responses are treated as false positives and removed. However, due to the error of manual annotation and local optimal learning of neural networks, many locations in the immediate neighborhood of corners or near the borders of the image may have slightly lower responses than those corners, and the threshold cannot effectively eliminate them. Non-maximum suppression (NMS), which has been used effectively in many object detection algorithms to solve the high overlap between the predicted bounding-boxes, can be adapted to our model with few modifications. Construct bounding-boxes (with area of $4\times 4$ pixels) centered around the remaining locations, then apply NMS on them based on the sorted response values, the satisfactory results can be got with the threshold at 0.5. In the result section we will show that NMS can eliminate the double detections without harming the ultimate detection accuracy.

**Clustering.** For pictures with complex scenes, there are many false positives that have very similar appearance to the corners, and therefore their response values are almost the same as those of corners, so that the techniques mentioned above can't distinguish them very well. Considering that the checkerboard has a very regular geometric property, while the false positives are distributed randomly and a little away from the checkerboard in the image, we can use the clustering algorithms to separate them. The k-means++ method is a widely used clustering technique that classes all points into certain clusters according to the minimized squared distance between points in the same cluster [24]. Here we apply this method to the remaining responses with $k=10$, then calculate the number of points $N_i (i=1…k.)$ in each cluster and eliminate points in the cluster with $N_i <= 2$.

## 3   The datasets

The datasets for training and testing of our model should be large enough with considering lens distortion, extreme poses and sensor noise. The training is performed on two image series: images captured by us directly and digitally augmented versions of these captured images.

For generating abundant training images, we used four types of checkerboards with $7\times 7, 6\times 9, 7\times 11, 9\times 9, 12\times 13$ inner corners as calibration patterns. Each pattern was placed under various circumstances to capture the datasets, with the background cluttered intentionally for simulating realistic calibration environments. In order to make our model become rotation (to some extent) and intensity invariant, here we rotated the original images by 90,180,270 degrees and reversed the intensities of half of those pictures randomly. The camera we used has little lens distortion and well capture conditions without much noise, so we artificially added both radial and tangential distortion as well as Gaussian noise as mentioned in MATE to multiply these pictures. Finally, all (a total of 8900 images) were converted to gray-scale images and resized into VGA resolution with 640×480 pixels (an optional operation) as the training dataset, as illustrated in Fig. 2. Among them 8000 images were selected randomly as the training dataset, and the rest were taken as the validation dataset.

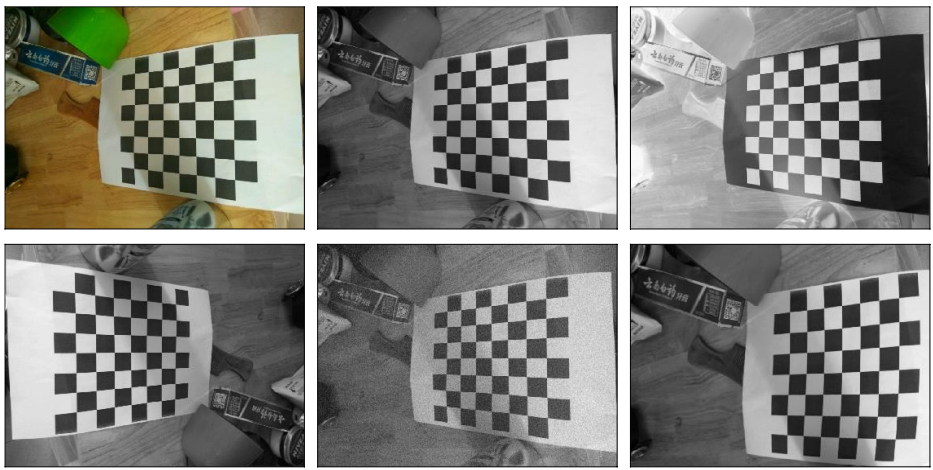

**Fig. 2.** Several sample images from the augmented data set. The top row shows the original image, the gray-scale image and the intensities inverted image. The second row shows the image rotated 180 degrees, the image with gaussian noise and the image with lens distortion.

The datasets for testing the generalization performance of our model consist two parts: the uEye and GoPro from ROCHADE [13]. The uEye dataset (with a resolution of 1280×1024 pixels) has slight lens distortion and serve to evaluate the robustness against perspective transforms and noise. The GpPro dataset (with a resolution of 4000×3000 pixels) is down-sampled to half-resolution and used to illustrate the robustness against lens distortion. These two datasets are shown in Fig. 3.

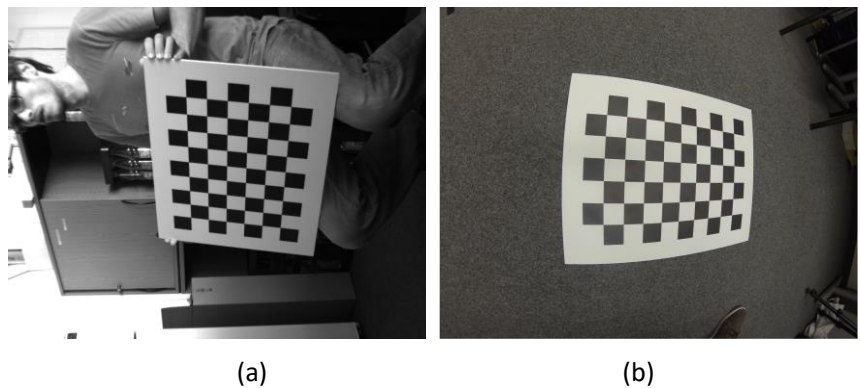

(a)          (b)

**Fig. 3.** Two examples from the uEye (a) and GoPro (b) datasets[13] with perspective transforms and lens distortion. The presented images are resized with nearest neighbor interpolation.

Checkerboard corner detection is a supervised learning task so the ground-truth corner locations should be obtained accurately. At first we annotated the four outer corners of the checkerboard manually, and then the interior corners were interpolated to converge locally to the saddle points. Finally, we checked and removed the wrong corners. After annotations we normalized per-pixel value to between 0 and 1 corresponding with the corner label mentioned in subsection 2.1, and so to some extent the response value can also be regarded as the probability to be a corner or not.

## 4    Experiments and Results

Two groups of experiments are presented to initiate a detailed study of the proposed model's performances. The first experiment is to test the learning ability of cross entropy as the loss function comparing with mean squared error (MSE) mentioned in MATE, as well as the feasibility of the learning rate with exponential decay. In order to make a quantitative comparison, we use the mean squared value (MSV) of the difference between the real label and the predicted value on all pixels in validation images.

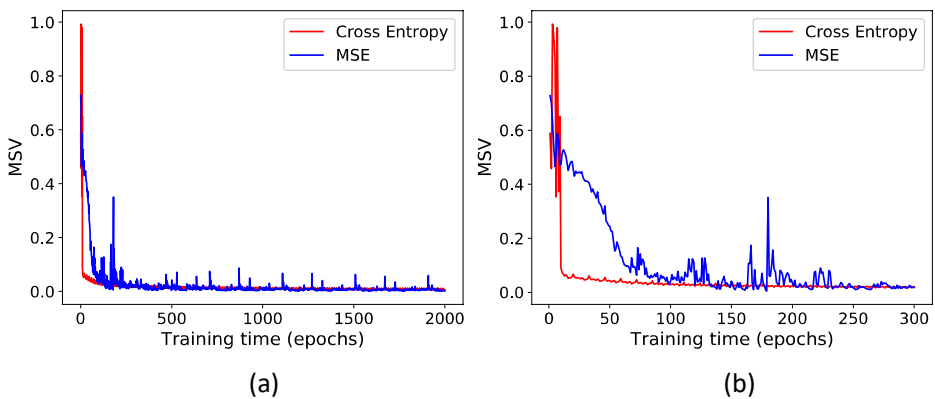

**Fig. 4.** MSV versus Training time (epochs) on the validation dataset. The neural network with cross entropy (red line) is equivalent to that with MSE (blue line), except that the initial learning rates were chosen independently to make training as good as possible.

We can see from Fig. 4(a) that cross entropy can get the similar result as MSE after 2000 epochs. But the neuron network with cross entropy drove down the cost rapidly while MSV of network with MSE started out much more slowly for the first 150 epochs, as shown in Fig. 4(b). The results comply with the view in [22] that mean squared error usually studies slowly when used with gradient-based optimization. Taken together, the cost value decreased rapidly at first but gradually slowed down without too many mutations, this is consistent with the design of the learning rate with exponential decay with the purpose to get a local optimal solution. The purpose of this experiment is not to show the learning ability of the whole model, but only to illustrate that the loss function and the learning rate with exponential decay presented are feasible techniques.

In the second group of experiments we performed several quantitative comparisons with the state-of-the-art methods MATE, ChESS, ROCHADE and OCamCalib with respect to the accuracy, the missed corner rate, double detection rate, and the number of false positives on the testing datasets. The distance between the detected corner and the closest ground truth corner of all images is calculated, and if the distance is less than five pixels, the detected point is regarded as a true corner. The accuracy denotes the average distance calculated above. The missed corner rate shows how many ground truths are detected as non-corners, and the double detection rate illustrates how many points close to each other are detected as the same corner. False positives show how many non-corner locations are detected as the corners on the whole dataset.

Public implementations of the last three algorithms can be used in this evaluation. However, considering that the training details and the hyper parameters of MATE are not available, we used the best published results shown in [16]. The results of these experiments are shown in Table 1 and Table 2.

**Table 1.** Results on the uEye dataset

| Method | Accuracy (px) | Missed Corners (%) | Double Detections (%) | False Positives |
|---|---|---|---|---|
| CCDN | 0.812 | 1.169 | 0.000 | 93 |
| MATE | 1.009 | 3.065 | 0.809 | 492 |
| ChESS | 0.946 | 3.398 | 0.000 | 11 |
| ROCHADE | 1.510 | 2.895 | 0.000 | 1 |
| OCamCalib | 0.319 | 0.000 | 0.000 | 0 |

**Table 2.** Results on the GoPro dataset

| Method | Accuracy (px) | Missed Corners (%) | Double Detections (%) | False Positives |
|---|---|---|---|---|
| CCDN | 0.576 | 0.907 | 0.000 | 0 |
| MATE | 0.835 | 4.566 | 4.556 | 389 |
| ChESS | 1.389 | 5.481 | 0.222 | 56 |
| ROCHADE | 1.807 | 5.593 | 0.000 | 3 |
| OCamCalib | 0.458 | 0.537 | 0.000 | 0 |

We can see above that the proposed model does not lose performance over these state-of-the-art methods, whether precision or recall. In terms of the accuracy, missed corners rate and double detection rate, our algorithm can get a much better detection result than MATE, as well as ChESS and ROCHADE. In particular, the accuracy can be better by using sub-pixel precision approaches [13, 25]. By adopting threshold related to maximum response, NMS and clustering, the number of false positives reduces significantly than MATE, and even maintains zero on the images with lens distortion. OCamCalib performs the best on the two datasets, but it requires the number of squares in checkerboard pattern in advance (that's why it didn't get any false positives and double detections), and it can only be used on the checkerboards with wide white border. So our model outperforms all tested methods on the generalization performance without any prior knowledge and is more adaptable to complex scenarios such as checkerboards with intensity reversal.

## 5     Conclusion

In this paper we have presented a novel checkerboard corner detection algorithm to find the inner corners of checkerboards with high robustness for most situations. This whole algorithm contains a checkerboard corner detection network (CCDN) and some post-processing techniques. CCDN is a fully convolutional neural network, which contains six convolutional layers and about 16000 parameters. It realizes a complex and efficient combination of detected features to select the checkerboard corner candidates efficiently with two improvements of loss function and learning rate. Threshold related to maximum response, non-maximum suppression and clustering are gathered as the post-processing to eliminate false positives in different cases. Quantitative comparisons on two different datasets in results show it outperforms the state-of-the-art methods including MATE, ChESS, ROCHADE, and OCamCalib without any prior knowledge of the checkerboard pattern. Thus it can be seen as a specific corner detector that is accurate, robust and suitable for automatic detection.

## Acknowledgement

This work is partially supported by the National Natural Science Foundation of China (Grant No. 51335004 and No. 91648203) and the International Science & Technology Cooperation Program of China (Grant No. 2016YFE0113600).